\documentclass[]{youtu}
\PassOptionsToPackage{table}{xcolor}
\usepackage{mathpazo}
\usepackage{hyperref}
\usepackage{booktabs}
\usepackage{array}
\usepackage{url}
\usepackage[utf8]{inputenc}
\usepackage[T1]{fontenc}
\usepackage{wrapfig}
\usepackage{amsfonts}
\usepackage{nicefrac}
\usepackage{microtype}
\usepackage[table]{xcolor}
\usepackage{bm}
\usepackage{makecell}
\usepackage{enumitem}
\usepackage{multirow}
\usepackage{colortbl}
\usepackage{pifont}
\usepackage[ruled,vlined]{algorithm2e}
\usepackage{algorithmic}

\usepackage{amsmath,amsthm,amssymb}
\usepackage{tcolorbox}
\usepackage{color}
\usepackage{graphicx}
\usepackage{subcaption}
\usepackage{caption}
\usepackage{natbib}
\usepackage{wrapfig}
\setcitestyle{numbers,square}
\hypersetup{
    pdftitle={Disentangling Long-Term Memory via Latent Neuro-Symbolic Reasoning},
    pdfauthor={Cai Ke et al.},
}

\title{Disentangling Long-Term Memory via \\ Latent Neuro-Symbolic Reasoning}
\author{Cai Ke\textsuperscript{$1$}, Xinghao Chen\textsuperscript{$1,2,3$}, Xiaoyu Shen\textsuperscript{$2$}, Keyu Chen\textsuperscript{$1$}, Siyu An\textsuperscript{$1 \dagger$},\\
Junnan Dong\textsuperscript{$1 \dagger$}, Ruifeng Xu\textsuperscript{$4 \dagger$}, Ruizhi Qiao\textsuperscript{$1$}, Xing Sun\textsuperscript{$1$}}
\affiliation{\textsuperscript{$1$}Tencent Youtu Lab\\\textsuperscript{$2$}Zhejiang Key Laboratory of Industrial Intelligence and Digital Twin, EIT, Ningbo, China\\\textsuperscript{$3$}Department of Computing, The Hong Kong Polytechnic University, Hong Kong, China\\\textsuperscript{$4$}Shenzhen Loop Area Institute, Shenzhen, China\\kecai@stu.hit.edu.cn}
\youtufinalcopy

\begin{document}

\abstract{Personalized agents are required to reason over long-term history interactions to infer both explicit preferences and implicit behavioral evidence. While early flat retrieval methods score memory fragments independently and neglect the distributed information, current structured memory frameworks rely on query-agnostic static graphs that fail to capture the context-dependent relations. Crucially, raw textual memories are inherently entangled and noisy, making fine-grained personalization and cross-session reasoning computationally prohibitive. To this end, we present \texttt{LGM}, a novel neuro-symbolic framework that shifts long-term memory disentanglement into a continuous latent space. Specifically, $(i)$ instead of persisting fixed graphs, we design a tailored latent graph construction with a sparse autoencoder. Subject to each query, it maps historical interactions into latent memory nodes and disentangles the memory traces into sparse concept activations, dynamically synthesizing query-aware relational edge weights. $(ii)$ A graph encoder then treats the query embedding as a conditioning preference to direct non-linear message passing across the task-specific latent subgraph. This yields a highly expressive memory representation for effective activations. Extensive experiments on long-term personalization benchmarks demonstrate that \texttt{LGM} significantly outperforms state-of-the-art baselines in capturing both explicit and implicit preferences while enabling personalized responses.}
\maketitle

\renewcommand{\thefootnote}{}
\footnotetext{$\dagger$ Corresponding authors.}

\section{Introduction} 
Personalized agents are increasingly urged to maintain coherent, long-horizon interactions with users~\cite{zhao2023survey,hu2025memory,zhang2025survey,ke2025flexibly,ke2026dynamic,dong2026deep,an2026toward}. Achieving seamless personalization requires reasoning over extensive historical interactions to infer both explicit preferences, e.g., specific constraints or instructions and implicit evidence, e.g., evolving habits, underlying decision patterns, and latent task dependencies. Existing approaches predominantly operate in the explicit textual space, falling into two main paradigms. Early efforts rely on flat retrieval, where historical turns or summary fragments are independently embedded and retrieved via vector similarity~\cite{lu2023memochat,zhong2024memorybank,xu2025mem,chhikara2025mem0,fang2026lightmem}. While efficient, flat retrieval scores memory fragments in isolation, inherently missing distributed, multi-hop evidence spread across distant sessions. To solve this, the current paradigm introduces structured memory frameworks, e.g., graph retrieval augmented-generation, abbreviated as GraphRAG, to model the complex relational dependencies~\cite{gutierrez2024hipporag,rezazadeh2025from,ke2026meta,dong2026youtu,dong2024cost,dong2024knowgpt}. While they heavily rely on query-agnostic, static graphs, the adjacency structure remains consequently fixed regardless of the incoming task, failing to infer dynamic, context-dependent relations that only surface under specific triggers buried in the queries.

% \begin{figure}[!t]
%   \centering
%   \includegraphics[width=0.85\columnwidth]{figures/1-Intro.pdf}
%   \caption{\texttt{LGM} dynamically aggregates scattered behavioral cues in the latent space to accurately answer user queries.}
%   \label{fig:intro}
% \end{figure}

\begin{wrapfigure}{r}{0.5\linewidth} 
\centering
\vspace{-2mm}
\includegraphics[width=\linewidth]{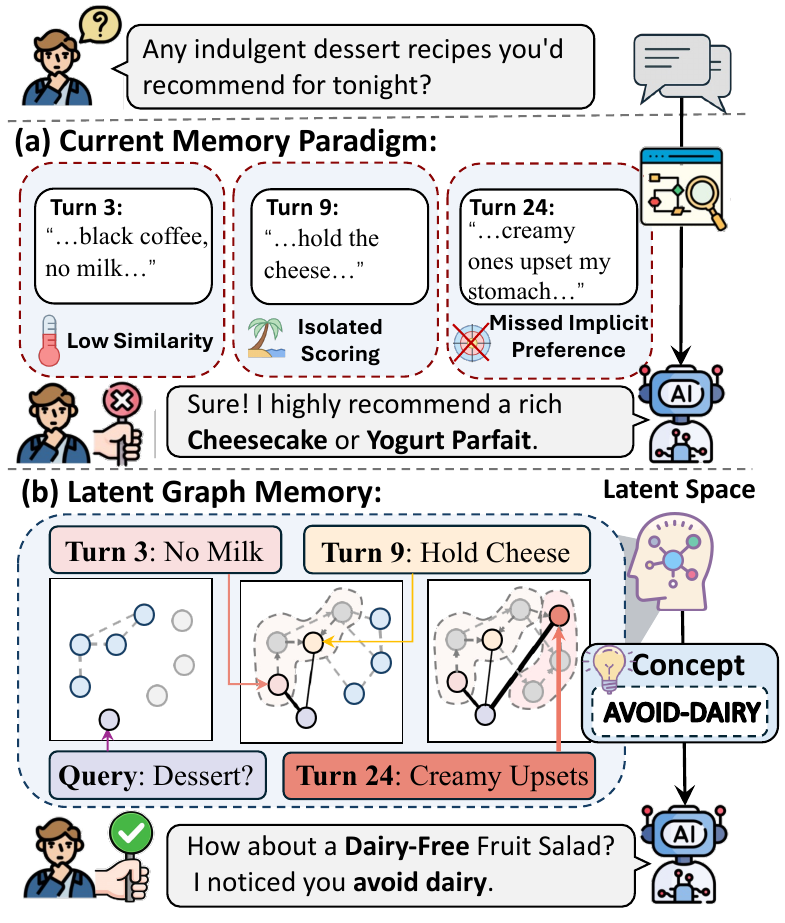}
\caption{\texttt{LGM} dynamically aggregates scattered behavioral cues in the latent space to accurately answer user queries.}
\label{fig:intro}
\vspace{-11mm}
\end{wrapfigure}

Nevertheless, raw textual memories are inherently entangled, noisy, and challenging to organize on demand. As illustrated in Figure~\ref{fig:intro}, traditional flat retrieval methods consider memory fragments in isolation, frequently missing distributed or implicit behavioral clues that are crucial for answering the query. Even when structured, a single interaction can simultaneously encode topic, intent, sentiment, and context—only a fraction of which is relevant to a specific request. Attempting fine-grained preference disentanglement and cross-session reasoning entirely within the token or prompt space introduces prohibitive computational overheads, often leading to ballooning KV caches, repeated language-model calls, and distraction from irrelevant text~\cite{chen2025reasoninglanguagecomprehensivesurvey}. This raises a central question:

\textbf{\emph{Can user long-term memory be disentangled and organized on demand, without repeatedly replaying the full textual history}}?

Answering this question is non-trivial, as it requires resolving three fundamental tensions. First, raw text and hidden states entangle explicit preferences with subtle habits. Disentangling these behavioral factors without losing long-tail evidence essential for personalization is difficult. Second, the relevance between historical events shifts with each query, making rigid text-level graphs too slow to rebuild and requiring the topology to adapt dynamically in the latent space. Finally, while reasoning in latent space avoids context window saturation, multi-step propagation risks information collapse, meaning the state must remain compact for fast generation while staying strictly faithful to historical evidence.

To address these challenges, we propose Latent Graph Memory, i.e., \texttt{LGM}, a neuro-symbolic framework that shifts long-term memory disentanglement and relational reasoning into a continuous latent space. Specifically, \((i)\) \textbf{Latent memory disentanglement}. \texttt{LGM} first aggregates multi-layer hidden states from the backbone language model and maps each historical interaction and the current query into a shared low-dimensional space. A sparse autoencoder further decomposes these dense representations into sparse concept activations, aligning direct preference statements and indirect behavioral cues within a shared concept space. \((ii)\) \textbf{Query-conditioned latent graph construction}. Instead of persisting a single global graph, \texttt{LGM} uses query/memory concept to construct a latent graph that co-activates to dynamically determine edge weights, allowing the same memory pool to form different connections under different cues. A relational graph neural network then propagates the query signal over the activated subgraph and combines distributed weak clues into a compact graph state with node-level relevance scores. \((iii)\) \textbf{Graph-Conditioned Reasoning}. \texttt{LGM} maps the dynamically synthesized latent graph into a compact set of memory prefix tokens, efficiently conditioning downstream generation without replaying raw texts. To ensure that continuous state propagation remains faithfully anchored to historical facts, we introduce an evidence-reconstruction objective that penalizes information loss and prevents latent space collapse. Consequently, the entire memory lifecycle from selection, association, aggregation, to utilization is unified into an end-to-end differentiable optimization process. Extensive experiments on PersonaMem~\cite{jiang2025know}, PrefEval~\cite{zhao2025do}, and the implicit-preference split of PersonaMem-v2~\cite{jiang2025personamem,dong2024clr} validate these advantages across Qwen2.5-7B, Gemma3-4B, and Qwen3-4B backbones. \texttt{LGM} achieves the highest average accuracy under both main backbones, outperforming the strongest baselines. The gains are most pronounced for implicit preferences and long-context settings, where successful responses require combining evidence scattered across sessions. \texttt{LGM} also occupies the low-cost frontier in language-model calls, token consumption, and time to first token. 

\textbf{Our contributions are summarized as follows:}

\begin{itemize}
    \item We formally formulate long-term personalization as query-aware memory disentanglement and relational reasoning in the latent space.
    \item We introduce \texttt{LGM}, an end-to-end neuro-symbolic framework that combines sparse concept decomposition, query-conditioned latent graph construction, relational message passing, and graph-conditioned generation.
    \item Extensive experiments are conducted across long-term personalization benchmarks and multiple language-model backbones. The results demonstrate consistent improvements in explicit and implicit preference, particularly under long contexts, while substantially reducing the computational overhead of textual memory replay.
\end{itemize}

\section{Related Work}

\paragraph{\textbf{Structured Retrieval-Augmented Generation.}}
A prominent line injects structured knowledge into Retrieval-Augmented
Generation (RAG) to better organize interaction history~\cite{sarthi2024raptor,rezazadeh2025from,edge2024local,guo-etal-2025-lightrag,gutierrez2024hipporag,dong2026youtu,dong2023hierarchy,dong2024knowgpt,dong2023active,dong2024cost,dong2024modality}.
Tree-based methods such as RAPTOR~\cite{sarthi2024raptor} and
MemTree~\cite{rezazadeh2025from} recursively cluster and summarize past texts
into hierarchies, while graph-based methods including
GraphRAG~\cite{edge2024local}, LightRAG~\cite{guo-etal-2025-lightrag},
HippoRAG~\cite{gutierrez2024hipporag}, and
Youtu-GraphRAG~\cite{dong2026youtu} link entities and relations for
multi-hop retrieval. \textbf{These methods commit to a global structure whose topology is fixed once
built; we defer structure to inference time, letting each query induce its own
latent topology so relevance is decided by the current cue rather than by a
pre-committed graph.}

\paragraph{\textbf{Agentic Memory Management.}}
Another line treats memory as an actively maintained store, compressing sessions
into summaries and user facts or letting agents decide when to write, merge, and
forget~\cite{zhong2024memorybank,chen2025compress,li-etal-2025-hello,wang2025recursively,chhikara2025mem0,xu2025mem,packer2023memgpt,kang-etal-2025-memory,fang2026lightmem}.
MemoryBank~\cite{zhong2024memorybank} and compressive-memory
approaches~\cite{chen2025compress,li-etal-2025-hello,wang2025recursively}
condense histories into user profiles; Mem0~\cite{chhikara2025mem0} and
A-Mem~\cite{xu2025mem} organize memories as interconnected notes; and
MemGPT~\cite{packer2023memgpt}, MemoryOS~\cite{kang-etal-2025-memory}, and
LightMem~\cite{fang2026lightmem} adopt OS-inspired hierarchies to schedule
storage. \textbf{These methods store memory as discrete textual records scored in
isolation, which caps evidence at the single-record level; we reason over memory
relationally in latent space, so scattered weak clues compose into evidence that
no record carries alone.}

\paragraph{\textbf{Reinforcement Learning for Memory.}}
Recent methods cast memory management as a sequential decision problem and
optimize read/write policies with reinforcement
learning (RL)~\cite{wang2025mem,yu2026memagent,xu2026learning,yan2026memory}.
MemAgent~\cite{yu2026memagent} maintains a fixed-size memory across long
contexts, while Mem-$\alpha$~\cite{wang2025mem} and
MemCoE~\cite{xu2026learning} learn unified policies over short- and long-term
memory. Memory-R1~\cite{yan2026memory} further trains an agent to manage memory. \textbf{Rather than optimizing discrete read/write policies over a fixed textual
memory with RL, we make memory itself differentiable,
turning memory management from an external control into an intrinsic learnable representation.}

\section{Problem Formulation}

We formalize personalized agent generation over long-term interaction histories. Let $\mathcal{H} = \{h_1, h_2, \dots, h_M\}$ denote a sequence of historical interactions spanning multi-session user-agent dialogues, and $q$ represent the current user query. Let $f_\theta(\cdot)$ be the backbone language model, $E(\cdot)$ its embedding lookup function, and $\mathbf{H}^{(l)} \in \mathbb{R}^{T \times d}$ the hidden representations extracted from layer $l \in \mathcal{L}$. The sequence $y = (y_1, \dots, y_N)$ denotes the generated target response.

\noindent\fbox{\parbox{\dimexpr\linewidth-2\fboxsep-2\fboxrule\relax}{Given an arbitrarily long user history $\mathcal{H}$ and an incoming query $q$, the objective is to generate a faithful, personalized response $y$ grounded in relevant historical evidence. The framework is required to process the raw history $\mathcal{H}$ to form a query-dependent memory state $\mathbf{M}_q$ that captures both explicit user preferences and context-scattered implicit behavior. Conditioned on both the current query embedding $E(q)$ and the constructed memory state $\mathbf{M}_q$, the backbone language model $f_\theta$ autoregressively predicts the probability distribution of the target response.}}

\section{Methodology}

\begin{figure*}[t]
    \centering
    \includegraphics[
        width=0.92\textwidth,
        keepaspectratio
    ]{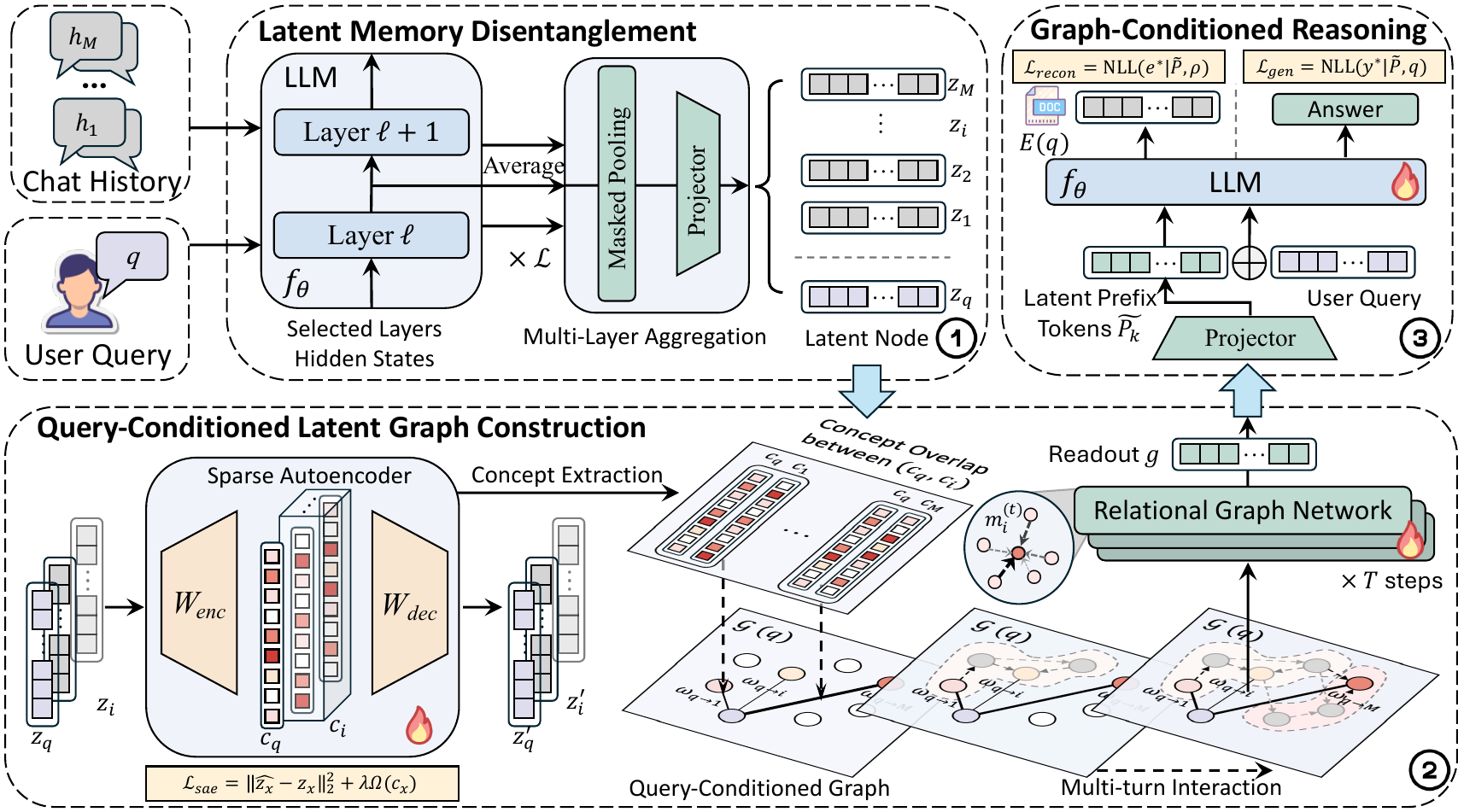}
    \caption{Overview of \texttt{LGM}: reasoning over memory on demand in the latent space.} 
    
    \label{fig:lgm_overview}
\end{figure*}

\subsection{Latent Memory Disentanglement}
\paragraph{Latent nodes and concepts.}
As shown in Figure~\ref{fig:lgm_overview}, rather than keeping memory as raw text, we represent every interaction and the
query in the model's own latent space. Each unit $x\in\mathcal{H}\cup\{q\}$ is
mapped to a \emph{latent node} $z_x\in\mathbb{R}^{d}$, giving a query node $z_q$
and memory nodes $\{z_i\}_{i=1}^{M}$. Because a single interaction simultaneously
encodes topic, intent, sentiment, and habit, a dense $z_x$ remains entangled. We
therefore attach to each latent a sparse concept code
\begin{equation}
  c_x = \mathcal{S}\big(\mathrm{ReLU}(W_{enc}\, z_x)\big),\qquad
  \hat{z}_x = W_{dec}\, c_x ,
  \label{eq:sae}
\end{equation}
produced by a sparse autoencoder (SAE)~\cite{yin2025constrain}, where
$\mathcal{S}(\cdot)$ suppresses weak activations and $\hat{z}_x$ is the
reconstruction. The sparse, non-negative code $c_x$ is the interpretable unit on
which explicit statements and implicit behavioral factors are aligned in a
common concept space.

\paragraph{Query-conditioned graph.}
We organize memory as a graph $\mathcal{G}(q)=(\mathcal{V},\mathcal{E},\omega)$
whose node set $\mathcal{V}=\{z_q\}\cup\{z_i\}$ contains the query and all
memories. Crucially, the edge weights $\omega$ are a \emph{function of the
query}: they are derived from the concept overlap between $c_q$ and $\{c_i\}$,
so the same memory pool induces a different topology for every incoming query.
The learning problem is thus to jointly (i) encode latent nodes and disentangle
them into concepts, (ii) build $\mathcal{G}(q)$ and propagate the query signal
into a compact state $g$, and (iii) condition $f_\theta$ on $g$ to reconstruct
evidence and generate $y$, all end to end.

We encode memory in the latent space so that behavioral signals not stated in
words are still preserved. Each unit $x$ is passed through $f_\theta$; the
selected-layer hidden states are averaged (multi-layer aggregation), reduced
along the sequence with masked pooling, and projected into the shared latent
space to obtain a latent node:
\begin{equation}
  z_x = \mathrm{LN}\!\Big(\sigma\big(W\cdot
        \mathrm{Pool}\big(\tfrac{1}{|\mathcal{L}|}\!\textstyle\sum_{\ell\in\mathcal{L}}
        f_\theta^{(\ell)}(x)\big)\big)\Big).
  \label{eq:encode}
\end{equation}
Averaging across layers is deliberate: lower layers carry lexical cues while
higher layers carry semantic and behavioral abstraction, and implicit
preferences typically live in the mixture rather than in any single layer.
Encoding is done in chunks so that memory grows in the \emph{number of latent
nodes} instead of a monolithic context window. Because the query is encoded by
the \emph{same} model (Eq.~\ref{eq:encode}), it lives in the identical space as
memory and can later act as a control signal instead of a mere retrieval key. Applying the SAE of Eq.~\eqref{eq:sae} then performs concept extraction,
disentangling each dense $z_x$ into a sparse code $c_x$. The SAE is trained to
reconstruct the latent while keeping the code sparse,
\begin{equation}
  \mathcal{L}_{sae}=\big\lVert \hat{z}_x - z_x \big\rVert_2^2
                    + \lambda\,\Omega(c_x),
  \label{eq:lsae}
\end{equation}
which is the step that separates \emph{what a memory is about} from \emph{how it
was said}, allowing a direct instruction and a faint recurring habit to be
matched through a shared concept even when their surface text is unrelated.

\subsection{Query-Conditioned Latent Graph Construction}
A dense similarity over latents would yield an entangled, query-agnostic graph.
Instead, we let the concept codes decide connectivity: the edge weight from the
query to a memory is their normalized concept overlap,
\begin{equation}
  \omega_{q\to i} = \phi\big(c_q,\, c_i\big)\in[0,1].
  \label{eq:edge}
\end{equation}
Since $c_q$ changes with the query, the query-conditioned graph
$\mathcal{G}(q)$ is rebuilt implicitly for every request: a memory that is
irrelevant under one cue can become strongly connected under another. We add
self-loops and typed, bidirectional query$\leftrightarrow$memory relations
$r_{ij}$ so the query can both gate memories and receive evidence back.

A relational graph network then runs $T$ steps of message passing over
$\mathcal{G}(q)$ across the multi-turn interaction. Starting from
$z_i^{(0)}=z_i$, each node aggregates relation-aware messages weighted by
Eq.~\eqref{eq:edge} and updates its state:
\begin{align}
  m_i^{(t)} &= \!\!\sum_{j\in\mathcal{N}(i)}\! \omega_{j\to i}\,
               \psi\big(z_i^{(t)}, z_j^{(t)}, r_{ij}\big), \\
  z_i^{(t+1)} &= \mathrm{Upd}\big(z_i^{(t)},\, m_i^{(t)}\big).
  \label{eq:mp}
\end{align}
A permutation-invariant readout then collapses the final states into a compact
memory state $g=\mathrm{Readout}(\{z_i^{(T)}\})$, and a scoring head produces a
per-node relevance $s_i=\langle z_i^{(T)},g\rangle$. This propagation is exactly
what turns several individually weak clues into one inferred implicit
preference where evidence scattered is combined along the query-activated edges, rather than being ranked by record.

\subsection{Graph-Conditioned Reasoning}
The state $g$ is a compressed, query-specific summary of the whole history. A
projector maps it into a small set of soft \emph{latent prefix} tokens
$\tilde{P}=\{\tilde{P}_k\}_{k=1}^{p}\in\mathbb{R}^{p\times d_{\text{lm}}}$, each
rescaled to the norm of ordinary token embeddings so that the prefix informs
rather than dominates attention. The prefix is prepended to the query embeddings
$E(q)$ and the response is generated autoregressively,
$y\sim f_\theta([\,\tilde{P};E(q)\,])$, so the long history influences generation
through a compact latent channel instead of being replayed as raw context.

\paragraph{Generative supervision.}
Because the target application is a conversational agent, we supervise the model
to \emph{generate the reference response text} $y^\star$ rather than to pick a
discrete option, which would collapse the task into option classification and
harm fluency:
\begin{equation}
  \mathcal{L}_{gen}
    = -\sum_{t}\log f_\theta\big(y^\star_t \mid y^\star_{<t},\,
      [\,\tilde{P};E(q)\,]\big).
  \label{eq:gen}
\end{equation}

\paragraph{Evidence reconstruction.}
A known failure mode of latent memory is that the prefix collapses into an
uninformative signal the model simply ignores~\cite{wei2025simcot,chen2026what}.
To keep $g$ anchored to facts, we reuse $f_\theta$ as a decoder and require the
\emph{same} prefix, under a fixed decoding instruction $\rho$, to reconstruct
the supporting evidence text $e^\star$:
\begin{equation}
  \mathcal{L}_{recon}
    = -\sum_{t}\log f_\theta\big(e^\star_t \mid e^\star_{<t},\,
      [\,\tilde{P};E(\rho)\,]\big).
  \label{eq:recon}
\end{equation}
This directly ties the latent channel to recoverable evidence and prevents
latent collapse; our ablation identifies $\mathcal{L}_{recon}$ as the single
most important term for making memory effective.

\paragraph{Overall objective.}
The full model is trained end to end, combining generation and reconstruction
with the SAE term:
\begin{equation}
  \mathcal{L} = \mathcal{L}_{gen}
    + \mathcal{L}_{recon}
    + \mathcal{L}_{sae}.
  \label{eq:total}
\end{equation}
Unlike retrieval methods that use the query only as a lookup key, \texttt{LGM} uses it to
control latent-node disentanglement, relation, and aggregation, so explicit
statements and distributed implicit clues are resolved within a single
mechanism.

\section{Experiments}
\subsection{Experimental Settings}

\paragraph{Datasets.}
We evaluate on two widely-used personalization benchmarks.
\textbf{PersonaMem}~\cite{jiang2025know} measures how well a model tracks and
updates user preferences across long, multi-session interaction histories, and
offers two context scales (32K and 128K tokens). \textbf{PrefEval}~\cite{zhao2025do}
probes preference following through \emph{explicit} and \emph{implicit}
preference queries embedded in long dialogues. To further stress implicit
personalization, we additionally evaluate on the implicit-preference split of
\textbf{PersonaMem-v2}~\cite{jiang2025personamem}, where the target preference is
never stated verbatim and must be inferred from scattered behavioral cues.

\paragraph{Model and Baselines.}
We build our method on three open backbones, \textbf{Qwen2.5-7B-Instruct}~\cite{yang2024qwen2}, \textbf{Gemma3-4B-it}~\cite{gemmateam2025gemma3technicalreport} and \textbf{Qwen3-4B-Instruct}~\cite{yang2025qwen3}, and compare against four families of
baselines. (i) \emph{Context / retrieval basics}: \textbf{Long Context}, which
directly feeds the raw history, and \textbf{Naive RAG}, which retrieves the
top-$K$ snippets from a vector store. (ii) \emph{Structured RAG}:
\textbf{GraphRAG}~\cite{edge2024local}, \textbf{LightRAG}~\cite{guo-etal-2025-lightrag}, \textbf{HippoRAG}~\cite{gutierrez2024hipporag},
and \textbf{Youtu-GraphRAG}~\cite{dong2026youtu}, which organize memory into graph
structures. (iii) \emph{Agentic memory management}: \textbf{MemoryBank}~\cite{zhong2024memorybank},
\textbf{Mem0}~\cite{chhikara2025mem0}, \textbf{A-Mem}~\cite{xu2025mem}, \textbf{MemoryOS}~\cite{kang-etal-2025-memory}, and
\textbf{LightMem}~\cite{fang2026lightmem}, which maintain and update an external memory bank.
(iv) \emph{RL-based memory}: \textbf{Mem-$\alpha$}~\cite{wang2025mem}, \textbf{MemAgent}~\cite{yu2026memagent}, \textbf{Memory-R1}~\cite{yan2026memory},
and \textbf{MemCoE}~\cite{xu2026learning}, which learn memory-update policies. For a fair
comparison, all baselines are run with the recommended settings from their
public codebases under the same backbones and data splits. All methods produce free-form
responses rather than selecting from predefined options.

\paragraph{Implementation Details.}
For each interaction and query, we read out hidden states from four evenly-spaced
intermediate layers ($\{8,16,24,32\}$), average them, and apply masked mean
pooling; a lightweight projector then maps the result into a compact latent
space ($d\!=\!32$). A differentiable sparse autoencoder (concept dimension $256$) disentangles each latent into sparse concepts whose
co-activation defines the query-conditioned edge weights. The relational graph
network uses a $256$-d hidden size and $T\!=\!4$ message-passing steps, and its
graph state is projected into $16$ latent prefix tokens, each rescaled to the
norm of ordinary token embeddings before being prepended to the query. The whole
system is trained end-to-end with AdamW~\cite{loshchilov2017decoupled}, using a learning rate of $2\text{e-}4$
for the graph modules and $1\text{e-}5$ for the backbone, weight decay $0.01$,
gradient clipping $1.0$, and $3$ epochs under bf16 mixed precision with gradient
checkpointing; the random seed is fixed for reproducibility. are conducted under the same environmental and hardware-level configurations. Long histories are encoded in chunks
(each segment truncated to $2048$ tokens) so that memory grows in the number of
latent nodes rather than in a context window.

\subsection{Main Results}
\paragraph{\textbf{Query-conditioned latent graphs turn scattered history into usable evidence, giving the largest gains exactly where implicit preferences must be inferred.}} Table~\ref{tab:main-results} reports results on PersonaMem and PrefEval under two backbones. Our method achieves the best average on both Qwen2.5-7B ($72.28$) and Gemma3-4B ($66.16$), improving over the strongest baseline MemCoE by $+11.20$ and $+11.70$. Three patterns stand out. First, flat pipelines (Long Context, Naive RAG) trail by a wide margin, confirming that independently ranked records fail to connect distributed evidence. Second, structured and agentic memories help but remain unstable across settings, since a pre-built topology cannot adapt to each query. Third, the gap is largest on the harder implicit split and on the long $128$K context, where our latent graph must combine several behavioral clues rather than copy a single stated fact. The consistent improvement across both backbones indicates that the gains come from the memory mechanism itself rather than a specific model.

\begin{figure*}[h]
    \centering

    \begin{minipage}[b]{0.74\textwidth}
        \centering
        \includegraphics[width=0.52\linewidth]{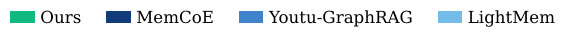}
    \end{minipage}\hfill
    \begin{minipage}[b]{0.24\textwidth} \centering \includegraphics[width=\linewidth]{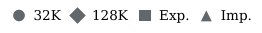} \end{minipage}

    \vspace{-0.5em}

    \begin{subfigure}[t]{0.24\textwidth}
        \centering
        \includegraphics[width=\linewidth]{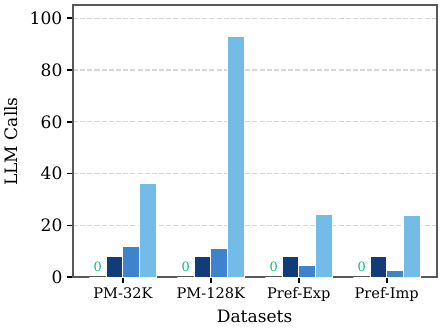}
        \caption{LLM calls}
        \label{fig:efficiency-calls}
    \end{subfigure}\hfill
    \begin{subfigure}[t]{0.24\textwidth}
        \centering
        \includegraphics[width=\linewidth]{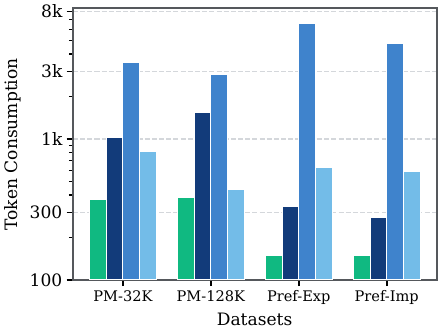}
        \caption{Token consumption}
        \label{fig:efficiency-tokens}
    \end{subfigure}\hfill
    \begin{subfigure}[t]{0.24\textwidth}
        \centering
        \includegraphics[width=\linewidth]{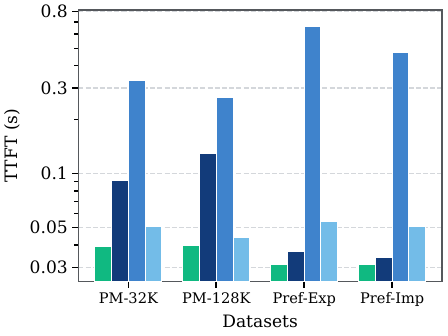}
        \caption{Time to first token}
        \label{fig:efficiency-ttft}
    \end{subfigure}\hfill
    \begin{subfigure}[t]{0.24\textwidth}
        \centering
        \includegraphics[width=\linewidth]{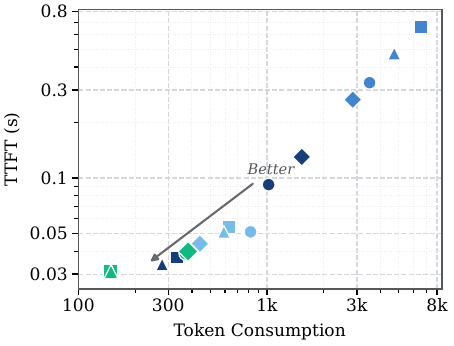}
        \caption{Token--latency trade-off}
        \label{fig:efficiency-pareto}
    \end{subfigure}

    \caption{
    Efficiency comparison across four datasets, where our method consistently achieves the lowest cost and latency.
    }
    \label{fig:efficiency-analysis}
\end{figure*}

\paragraph{\textbf{Operating in latent space instead of replaying text, our method reaches the accuracy frontier while sitting at the low-cost corner of every efficiency axis.}} Figure~\ref{fig:efficiency-analysis} compares efficiency against representative memory methods. Because history is compressed into a compact latent state rather than re-read as raw context, our method uses the fewest LLM calls (a) and the least token consumption (b), avoiding the repeated prompting and long contexts that inflate text-based memories. This directly lowers time to first token (c), keeping latency stable as the context grows to $128$K. The token--latency trade-off (d) summarizes the picture: our method occupies the bottom-left frontier across all datasets, delivering the strongest accuracy in Table~\ref{tab:main-results} at the lowest cost, whereas graph-construction and hierarchical-scheduling baselines pay a heavy overhead for weaker results.

\definecolor{deltaLow}{HTML}{5A9A5A}
\definecolor{deltaMid}{HTML}{2F7D32}
\definecolor{deltaHigh}{HTML}{145A20}

\newcommand{\mcsource}[1]{#1$^{*}$}
\newcommand{\deltalow}[1]{\textcolor{deltaLow}{+#1}}
\newcommand{\deltamid}[1]{\textcolor{deltaMid}{+#1}}
\newcommand{\deltahigh}[1]{\textcolor{deltaHigh}{+#1}}

\begin{table*}[!t]
\centering
\small
\setlength{\tabcolsep}{1.5pt}
\renewcommand{\arraystretch}{1.15}
\begin{tabular}{l cccccc cccccc}
\toprule
& \multicolumn{6}{c}{\textbf{Qwen2.5-7B-Instruct}}
& \multicolumn{6}{c}{\textbf{Gemma3-4B-it}} \\
\cmidrule(lr){2-7} \cmidrule(lr){8-13}
& \multicolumn{2}{c}{PersonaMem}
& \multicolumn{2}{c}{PrefEval}
& \multirow{2}{*}{Avg}
& \multirow{2}{*}{$\Delta$}
& \multicolumn{2}{c}{PersonaMem}
& \multicolumn{2}{c}{PrefEval}
& \multirow{2}{*}{Avg}
& \multirow{2}{*}{$\Delta$} \\
\cmidrule(lr){2-3}
\cmidrule(lr){4-5}
\cmidrule(lr){8-9}
\cmidrule(lr){10-11}
\multirow{-3}{*}{Method}
& 32K & 128K & Explicit & Implicit & &
& 32K & 128K & Explicit & Implicit & & \\
\midrule

Long Context
& \mcsource{34.36}
& \mcsource{25.05}
& \mcsource{31.70}
& \mcsource{30.80}
& 30.48
& \deltahigh{41.80}
& 20.69
& 24.63
& 48.10
& 39.80
& 33.31
& \deltahigh{32.85} \\

Naive RAG
& \mcsource{48.67}
& \mcsource{38.90}
& \mcsource{47.80}
& \mcsource{32.40}
& 41.94
& \deltahigh{30.34}
& 27.59
& 31.62
& 51.10
& 38.30
& 37.15
& \deltahigh{29.01} \\

\rowcolor{gray!15}
\multicolumn{13}{c}{
  \textit{Structured Retrieval-Augmented Generation}
} \\

GraphRAG {\scriptsize(arXiv'24, 34.9K stars)}
& \underline{63.79}
& 38.24
& 51.20
& 37.70
& 47.73
& \deltahigh{24.55}
& 51.72
& 40.44
& \underline{58.00}
& 69.70
& \underline{54.97}
& \deltalow{11.19} \\

HippoRAG {\scriptsize(NIPS'24)}
& \underline{63.79}
& \underline{52.38}
& 67.80
& 45.20
& 57.29
& \deltalow{14.99}
& 53.45
& 48.90
& 51.70
& 41.90
& 48.99
& \deltamid{17.17} \\

LightRAG {\scriptsize(EMNLP'25)}
& 51.72
& 38.60
& 51.10
& 38.00
& 44.86
& \deltahigh{27.42}
& 50.00
& 37.50
& 41.90
& 35.10
& 41.13
& \deltahigh{25.03} \\

Youtu-GraphRAG {\scriptsize(ICLR'26)}
& \underline{63.79}
& 38.60
& 57.00
& 42.20
& 50.40
& \deltamid{21.88}
& 46.55
& 41.91
& 42.90
& 35.00
& 41.59
& \deltamid{24.57} \\

\rowcolor{gray!15}
\multicolumn{13}{c}{
  \textit{Agentic Memory Management}
} \\

MemoryBank {\scriptsize(AAAI'24)}
& 58.62
& 45.22
& \underline{75.50}
& 47.90
& 56.81
& \deltamid{15.47}
& 51.72
& 48.16
& 55.80
& 45.20
& 50.22
& \deltamid{15.94} \\

Mem0 {\scriptsize(arXiv'25, 61.8K stars)}
& \mcsource{48.53}
& \mcsource{39.67}
& \mcsource{57.60}
& \mcsource{46.40}
& 48.05
& \deltamid{24.23}
& 53.45
& \underline{53.45}
& 40.81
& 36.00
& 45.93
& \deltahigh{20.23} \\

A-Mem {\scriptsize(NIPS'25)}
& \mcsource{48.26}
& \mcsource{38.22}
& \mcsource{62.30}
& \mcsource{52.80}
& 50.40
& \deltamid{21.88}
& 51.72
& 41.18
& 50.50
& 39.40
& 45.70
& \deltamid{20.46} \\

MemoryOS {\scriptsize(EMNLP'25)}
& 62.07
& 37.50
& 67.20
& 59.10
& 56.47
& \deltamid{15.81}
& 53.45
& 40.07
& 59.20
& 49.40
& 50.53
& \deltamid{15.63} \\

LightMem {\scriptsize(ICLR'26)}
& \mcsource{50.72}
& \mcsource{39.93}
& \mcsource{64.20}
& \mcsource{54.80}
& 52.41
& \deltamid{19.87}
& 48.28
& 39.71
& \underline{58.00}
& 65.20
& 52.80
& \deltalow{13.36} \\

\rowcolor{gray!15}
\multicolumn{13}{c}{
  \textit{Reinforcement Learning for Memory}
} \\

Mem-$\alpha$ {\scriptsize(arXiv'25)}
& \mcsource{53.37}
& \mcsource{42.86}
& 45.30
& 78.80
& 55.08
& \deltamid{17.20}
& 37.93
& 38.24
& 40.40
& \underline{74.70}
& 47.82
& \deltamid{18.34} \\

MemAgent {\scriptsize(ICLR'26)}
& \mcsource{53.58}
& \mcsource{43.59}
& 45.00
& 80.80
& 55.74
& \deltamid{16.54}
& \underline{56.90}
& 43.75
& 43.90
& 69.70
& 53.56
& \deltalow{12.60} \\

Memory-R1 {\scriptsize(ACL'26)}
& 54.79
& 43.12
& 47.60
& 79.40
& 56.23
& \deltamid{16.05}
& 55.17
& 43.01
& 44.60
& 69.90
& 53.17
& \deltalow{12.99} \\

MemCoE {\scriptsize(ACL'26)}
& \mcsource{57.06}
& \mcsource{47.24}
& 58.90
& \underline{81.10}
& \underline{61.08}
& \deltalow{11.20}
& 56.62
& 44.12
& 46.80
& 70.30
& 54.46
& \deltalow{11.70} \\

\midrule
\rowcolor{green!10}
\textbf{LGM (Ours)}
& \textbf{64.23}
& \textbf{65.00}
& \textbf{78.50}
& \textbf{81.40}
& \textbf{72.28}
& --
& \textbf{62.31}
& \textbf{66.34}
& \textbf{59.40}
& \textbf{76.60}
& \textbf{66.16}
& -- \\
\bottomrule
\end{tabular}

\caption{
Accuracy (\%) on PersonaMem and PrefEval under
Qwen2.5-7B and Gemma3-4B.
$\Delta$ denotes the absolute improvement of \textbf{Ours} over each
method's Avg, with darker green indicating a larger gain.
The best results are shown in \textbf{bold}, while the second-best
results are \underline{underlined}.
Values marked with $^{*}$ are taken from the MemCoE~\cite{xu2026learning}.
}
\label{tab:main-results}
\end{table*}

% \begin{figure}[!t]
%     \centering
%     \includegraphics[width=0.3\linewidth]{figures/ablation_gap_legend.pdf}\\[2pt]
%     \begin{subfigure}[t]{0.49\linewidth}
%         \centering
%         \includegraphics[width=\linewidth]{figures/ablation_gap_qwen.pdf}
%         \caption{Qwen2.5-7B}
%         \label{fig:ablation-qwen}
%     \end{subfigure}
%     \hfill
%     \begin{subfigure}[t]{0.49\linewidth}
%         \centering
%         \includegraphics[width=\linewidth]{figures/ablation_gap_gemma.pdf}
%         \caption{Gemma3-4B}
%         \label{fig:ablation-gemma}
%     \end{subfigure}
%     \caption{Ablation Study on PrefEval and PersonaMem.}
%     \label{fig:ablation}
% \end{figure}

\begin{figure}[!t]
    \centering
    \includegraphics[width=0.6\linewidth]{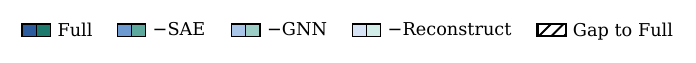}\\[2pt]
    \begin{subfigure}[t]{0.32\linewidth}
        \centering
        \includegraphics[width=\linewidth]{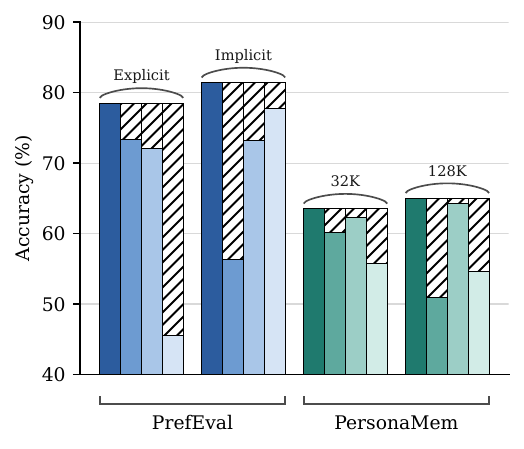}
        \caption{Qwen2.5-7B}
        \label{fig:ablation-qwen}
    \end{subfigure}%
    \hspace{0.01\linewidth}%
    \begin{subfigure}[t]{0.32\linewidth}
        \centering
        \includegraphics[width=\linewidth]{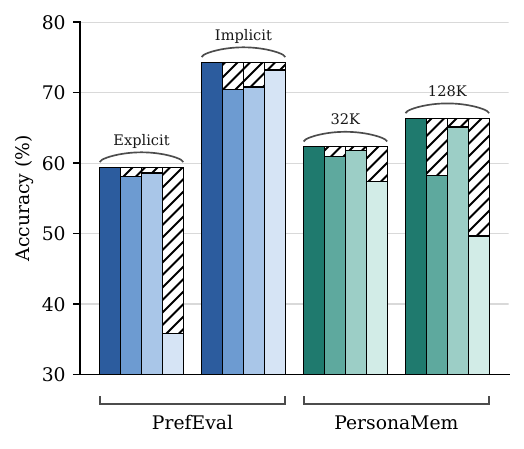}
        \caption{Gemma3-4B}
        \label{fig:ablation-gemma}
    \end{subfigure}
    \caption{Ablation Study on PrefEval and PersonaMem.}
    \label{fig:ablation}
\end{figure}

\subsection{Ablation Study}
\textbf{Each individual module proves clearly effective for both explicit and implicit preferences.} Figure~\ref{fig:ablation} carefully ablates three key components of our framework: the sparse autoencoder ($-$SAE), the graph propagation ($-$GNN), and the evidence-reconstruction objective ($-$Reconstruct); the hatched caps clearly mark the remaining gap up to the full model. Removing any single component consistently hurts overall performance across both backbones, further confirming that each one is truly necessary. Among them, $-$Reconstruct causes by far the largest drop, especially on the harder implicit preferences, where the latent prefix would otherwise collapse into an uninformative signal that the model ignores; grounding it to supporting evidence is exactly what keeps the aggregated memory usable. $-$SAE and $-$GNN also clearly degrade the results, indicating that disentangling hidden states into sparse concepts and propagating the query over the induced topology are both needed to connect distributed clues rather than treat memories in isolation.

\begin{wraptable}{r}{0.5\linewidth} 
\centering
\renewcommand{\arraystretch}{1.2}
\scriptsize
\setlength{\tabcolsep}{1.5mm}{
\begin{tabular}{lcc}
\toprule
Method & 32K & 128K \\
\midrule
\rowcolor{gray!15}
\multicolumn{3}{c}{\textit{Closed-Source Models}} \\
GPT-5-Chat$^{\dagger}$     & 45.6 & 41.4 \\
GPT-5-mini$^{\dagger}$     & 48.7 & \underline{44.1} \\
GPT-5-nano$^{\dagger}$     & --   & 33.9 \\
GPT-4.1$^{\dagger}$        & --   & 38.2 \\
GPT-4.1-mini$^{\dagger}$   & --   & 37.5 \\
o4-mini$^{\dagger}$        & --   & 38.9 \\
\rowcolor{gray!15}
\multicolumn{3}{c}{\textit{Open-Source \& Agentic Methods}} \\
Qwen3-4B-Base$^{\dagger}$  & 30.5 & --   \\
Qwen3-4B-SFT$^{\dagger}$   & 35.0 & --   \\
Qwen3-4B-GRPO$^{\dagger}$  & 35.6 & --   \\
Agentic Memory (SOTA)$^{\dagger}$ & \underline{55.2} & --   \\
\midrule
\rowcolor{green!10}
\textbf{Ours (Qwen3-4B)}  & \textbf{62.9}{\scriptsize$^{(+7.7)}$} & \textbf{51.9}{\scriptsize$^{(+7.8)}$} \\
\bottomrule
\end{tabular}}
\caption{Accuracy (\%) on PersonaMem-V2 under the 32K and 128K dialogue history settings. Methods marked with $^{\dagger}$ are taken from the original benchmark~\cite{jiang2025personamem}; ``--'' denotes settings not reported therein.
}
\label{tab:personamem-v2}
\end{wraptable}

\begin{wrapfigure}{r}{0.5\linewidth} 
\vspace{-10cm}
  \centering
  \includegraphics[width=\linewidth]{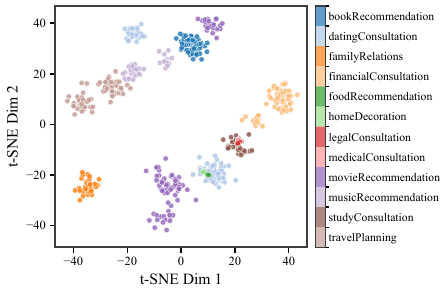}
  \caption{t-SNE visualization of graph states.}
  \label{fig:tsne-states}
\end{wrapfigure}

\subsection{Analysis on Inferring Implicit Preferences}
\paragraph{\textbf{LGM outperforms strong proprietary LLMs on inferring implicit preferences, despite using only a 4B setting.}} We further evaluate on a much harder implicit-preference setting, where user personas are revealed through behavior and choices rather than stated explicitly, so answers must be inferred by combining distributed clues. Using the same Qwen3-4B backbone against official benchmark numbers, our method reaches $62.9$ at $32$K and $51.9$ at $128$K, the best in both settings (Table~\ref{tab:personamem-v2}). Despite using only a $4$B open backbone, we clearly surpass strong proprietary models such as the GPT-5 and GPT-4.1 families (e.g., GPT-5-Chat $45.6$ / $41.4$); among methods on the same Qwen3-4B, we also exceed prompting- and RL-tuned variants (Base/SFT/GRPO) and their agentic-memory ($55.2$) by $+7.7$. This confirms the gains come from query-conditioned latent reasoning rather than model scale, and that \texttt{LGM} is most effective precisely where preferences must be inferred.

\subsection{Visualization of Graph States}
\paragraph{\textbf{The graph states show that our memory encodes semantically meaningful structure.}} Figure~\ref{fig:tsne-states} shows a t-SNE projection of the graph states produced on PersonaMem-32K with Gemma3-4B, colored by scenario. The states separate into well-defined clusters that align with the underlying scenarios, indicating that the query-conditioned graph aggregates history into a compact state that preserves user- and topic-level semantics rather than a noisy mixture. Semantically related scenarios also lie closer in the space, suggesting the memory captures meaningful relations across topics. This organized structure supports accurate memory selection and helps explain the gains observed in the main results.

\subsection{Effect of Latent Token Number}
\begin{figure}[!t]
    \centering
    \includegraphics[width=0.6\linewidth]{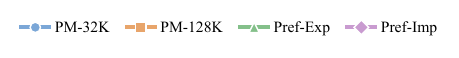}\\[1pt]
    \begin{subfigure}{0.32\linewidth}
        \centering
        \includegraphics[width=\linewidth]{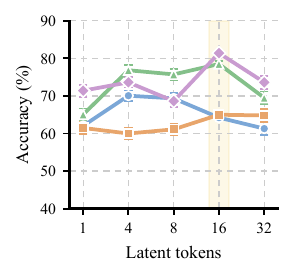}
        \caption{Qwen2.5-7B}\label{fig:latent-qwen}
    \end{subfigure}%
    \hspace{0.01\linewidth}%
    \begin{subfigure}{0.32\linewidth}
        \centering
        \includegraphics[width=\linewidth]{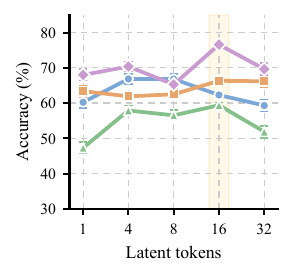}
        \caption{Gemma3-4B}\label{fig:latent-gemma}
    \end{subfigure}
    \caption{Effect of latent token number.}
    \label{fig:latent-tokens}
\end{figure}

\paragraph{\textbf{A small latent budget already conveys memory effectively.}} Figure~\ref{fig:latent-tokens} varies the number of latent prefix tokens from $1$ to $32$. Accuracy rises quickly with just a few tokens and peaks around $16$, then gradually saturates or slightly declines with more tokens, so we adopt $16$ as the default. This trend is consistent across both backbones and all four settings, clearly showing that a compact latent channel is sufficient to carry the aggregated memory: too few tokens underfit the query-conditioned state, while too many add redundant capacity that dilutes the signal and offers no further gain.

\subsection{Case Study: Explicit vs.\ Implicit Preferences}
\begin{wraptable}{r}{0.5\linewidth}
\vspace{-2mm}
\centering
\small
\setlength{\tabcolsep}{5pt}
\renewcommand{\arraystretch}{1.15}
\definecolor{okgreen}{HTML}{1B7F3B}
\definecolor{badred}{HTML}{C0392B}
\definecolor{oursblue}{HTML}{1F4E9C}
\newcommand{\cmark}{\textcolor{okgreen}{\ding{51}}}
\newcommand{\xmark}{\textcolor{badred}{\ding{55}}}
\begin{tabular}{@{}l cc cc@{}}
\toprule
\multirow{2}{*}{\textbf{Method}}
 & \multicolumn{2}{c}{\textbf{Case A}} & \multicolumn{2}{c}{\textbf{Case B}} \\
\cmidrule(lr){2-3}\cmidrule(lr){4-5}
 & Explicit & Implicit & Explicit & Implicit \\
\midrule
MemCoE          & \cmark & \xmark & \cmark & \xmark \\
Youtu-GraphRAG  & \cmark & \xmark & \cmark & \xmark \\
LightMem        & \cmark & \xmark & \cmark & \xmark \\
\rowcolor{green!10}\textbf{Ours}
 & \cmark & \cmark & \cmark & \cmark \\
\midrule
\multicolumn{5}{@{}p{0.95\linewidth}@{}}{\footnotesize
\textbf{Case A} (health constraint) — \textit{Q:} ``recommend an indulgent dessert''.
\textbf{Explicit:} ``I'm lactose intolerant and \textcolor{oursblue}{avoid dairy} entirely.''
\textbf{Implicit:} ``the salad fits my \textcolor{oursblue}{\emph{dietary needs}}, the others don't.''
\;$\Rightarrow$ correct: dairy-free salad (the only option without dairy;
distractors: yogurt / cream-cheese / grilled-cheese).}\\[2pt]
\multicolumn{5}{@{}p{0.95\linewidth}@{}}{\footnotesize
\textbf{Case B} (taste aversion) — \textit{Q:} ``suggest top-rated strategy games''.
\textbf{Explicit:} ``I prefer games without \textcolor{oursblue}{historical war settings}.''
\textbf{Implicit:} ``the \textcolor{oursblue}{\emph{war settings}} of the others don't appeal to me.''
\;$\Rightarrow$ correct: city-building sim (the only non-war title;
distractors: WWII FPS / Roman RTS / war RPG).}\\
\bottomrule
\end{tabular}
\caption{Case study on PrefEval (Gemma3-4B).}

\label{tab:case_study}
\vspace{-8mm}
\end{wraptable}

\textbf{Baselines collapse once a preference is only implied, whereas our
concept-level alignment still honors it.}
Table~\ref{tab:case_study} reports two requests where the same preference appears
both as an explicit statement and as an indirect behavioral cue. The explicit
form overlaps lexically with the query, so every method selects the correct
option. The gap emerges in the implicit setting: the relevant signal shares no
surface tokens with the query, so text-retrieval baselines find no match and
fall back to a preference-violating distractor (a dairy dessert in Case~A, a
war-themed title in Case~B). \texttt{LGM} instead maps both the explicit
statement and the implicit cue onto the \emph{same} sparse concepts and links
them to the query through the latent graph, selecting the
preference-consistent option regardless of how the preference is voiced.

\section{Conclusion}
In this work, we recast long-term personalization as query-aware memory
disentanglement and relational reasoning in a continuous latent space. Our
end-to-end framework, \texttt{LGM}, encodes interactions into latent nodes,
disentangles them into sparse concepts, induces a query-conditioned latent
graph, and propagates the query into a compact state that conditions generation,
kept faithful by an evidence-reconstruction objective. Across three benchmarks
and multiple backbones, \texttt{LGM} consistently improves both explicit and
implicit preferences, with the largest gains under long contexts, while
remaining on the low-cost frontier in calls, tokens, and latency. This shows
that organizing memory on demand in the latent space, rather than replaying raw
history, is a promising and practical path toward personalized agents.

\bibliography{references}
\appendix
\section*{Appendix}

\noindent This appendix provides additional details omitted from the main paper
due to space constraints. Appendix~\ref{app:datasets} describes the
personalization benchmarks used for evaluation; Appendix~\ref{app:baselines}
details every compared baseline together with the protocol we follow for a fair
comparison; Appendix~\ref{app:algorithm} gives the full algorithmic description
of \texttt{LGM}; and Appendix~\ref{app:hyper} lists the complete set of
implementation and hyperparameter settings.

% =====================================================================
\section{Dataset Details}
\label{app:datasets}
% =====================================================================

We evaluate \texttt{LGM} on three long-term personalization benchmarks:
\textbf{PersonaMem}~\cite{jiang2025know}, \textbf{PrefEval}~\cite{zhao2025do},
and the harder implicit split of
\textbf{PersonaMem-v2}~\cite{jiang2025personamem}. All three are designed to
probe whether a model can track a user across long, multi-session histories and
respond consistently with the user's \emph{explicit} preferences (directly
stated) and \emph{implicit} preferences (only inferable from scattered
behavioral cues). Table~\ref{tab:dataset-stats} summarizes their key
characteristics; we describe each benchmark below and then detail the shared
evaluation protocol.

\begin{table}[t]
\centering
\small
\setlength{\tabcolsep}{4pt}
\renewcommand{\arraystretch}{1.2}
\begin{tabular}{lccc}
\toprule
\textbf{Property} & \textbf{PersonaMem} & \textbf{PrefEval} & \textbf{PersonaMem-v2} \\
\midrule
Context scale     & 32K / 128K   & long dialogue & 32K / 128K \\
Sessions          & up to 60     & multi-turn    & multi-session \\
Explicit pref.    & \checkmark   & \checkmark    & \checkmark \\
Implicit pref.    & \checkmark   & \checkmark    & \checkmark~(focus) \\
Answer form       & free-form    & free-form     & free-form \\
Primary metric    & Accuracy     & Accuracy      & Accuracy \\
\bottomrule
\end{tabular}
\caption{Summary of the three personalization benchmarks used in our
experiments. ``Implicit pref.~(focus)'' indicates that the split is
specifically constructed so that the target preference is never stated
verbatim.}
\label{tab:dataset-stats}
\end{table}

\paragraph{PersonaMem.}
PersonaMem~\cite{jiang2025know} is a large-scale benchmark for \emph{dynamic
user profiling} that measures how well a model tracks and updates user
preferences across long, multi-session interaction histories. It comprises over
$180$ simulated user--LLM interaction histories, each with up to $60$ multi-turn
sessions (reaching roughly $1$M tokens in the largest configurations) that span
$15$ diverse personalized task scenarios. Each instance consists of an extended
user--agent history in which the user reveals, refines, and occasionally revises
preferences over time, followed by a query whose correct answer depends on the
\emph{most up-to-date} state of the user profile. The benchmark deliberately
stresses two abilities: (i) \emph{persistence}, i.e., remembering a preference
stated many turns ago, and (ii) \emph{updating}, i.e., overriding an earlier
preference when the user later changes their mind. To decouple the memory
mechanism from the underlying context window, we evaluate at two context scales,
\textbf{32K} and \textbf{128K} tokens; the 128K setting is substantially harder
because the relevant evidence is diluted across a much longer history and is
prone to the ``lost in the middle'' effect~\cite{liu2024lost}. We report
accuracy separately for both scales.

\paragraph{PrefEval.}
PrefEval~\cite{zhao2025do} evaluates \emph{personalized preference following} in
a long-context conversational setting. It contains $3{,}000$ manually curated
user preferences spanning $20$ everyday topics, and probes whether a model can
\emph{infer, memorize, and adhere to} a preference expressed earlier in a long
dialogue when asked a related downstream question. Its distinctive feature is
that the \emph{same} preference is provided under two expression modes. In the
\textbf{explicit} setting, the preference is stated directly (e.g., ``I am
lactose intolerant and avoid dairy''), so it overlaps lexically with the query
and is easy to match by surface retrieval. In the \textbf{implicit} setting, the
preference is only conveyed through an indirect behavioral cue (e.g., ``the
salad fits my dietary needs, the others don't''), sharing no surface tokens with
the query and thus requiring the model to \emph{infer} the underlying preference
rather than copy it. This explicit/implicit contrast makes PrefEval a direct
probe of whether a memory mechanism can align differently-voiced expressions of
the same preference, which is precisely the setting our concept-level alignment
targets. We report accuracy on the explicit and implicit splits separately.

\paragraph{PersonaMem-v2.}
To further stress implicit personalization, we additionally evaluate on the
implicit-preference split of PersonaMem-v2~\cite{jiang2025personamem}. Unlike the
original PersonaMem, PersonaMem-v2 is explicitly built around \emph{implicit
user personas}: it covers $1{,}000$ comprehensive user personas and over
$20{,}000$ preferences across $300$+ scenarios, and its target preference is
\emph{never stated verbatim} but must be reconstructed by combining scattered
behavioral evidence (choices, reactions, and habits) observed across the
dialogue history. This makes it the most challenging of the three benchmarks, as
it eliminates any lexical shortcut between the query and the supporting evidence.
Following the original benchmark, we evaluate at the \textbf{32K} and
\textbf{128K} dialogue-history settings and compare against the officially
reported numbers for both proprietary models (the GPT-5 and GPT-4.1 families) and
open-source / agentic baselines, using the same Qwen3-4B backbone.

\paragraph{Evaluation Protocol.}
For all three benchmarks, every method (including all baselines) produces a
\emph{free-form} natural-language response rather than selecting from predefined
options; this avoids collapsing the task into option classification and better
reflects the conversational agent setting. Following the benchmark conventions,
we score each response for accuracy against the reference answer /
preference-consistent target. To ensure comparability across methods, we fix the
backbone, data splits, and context scales, and evaluate all systems under
identical conditions.

% =====================================================================
\section{Baseline Details}
\label{app:baselines}
% =====================================================================

We compare \texttt{LGM} against four families of baselines that span the main
paradigms for equipping language models with long-term memory. Below we describe
each method and the rationale for its inclusion. For a fair comparison, all
baselines are run with the recommended settings from their public codebases,
under the \emph{same} backbones (Qwen2.5-7B-Instruct~\cite{yang2024qwen2},
Gemma3-4B-it~\cite{gemmateam2025gemma3technicalreport}, and
Qwen3-4B-Instruct~\cite{yang2025qwen3}) and the same data splits used by our
method. All methods generate free-form responses.

\paragraph{(i) Context / Retrieval Basics.}
These two baselines establish the lower and upper ends of the naive spectrum.
\begin{itemize}
\item \textbf{Long Context} directly feeds the entire raw interaction history
into the backbone's context window without any memory mechanism. It represents
the ``no-memory'' baseline on raw context usage, and its degradation at 128K
illustrates the ``lost in the middle'' problem~\cite{liu2024lost}.
\item \textbf{Naive RAG} builds a vector store over the history and retrieves
the top-$K$ most similar snippets for each query, appending them to the prompt.
It represents standard flat retrieval, where memory fragments are scored
\emph{independently} and cannot connect distributed evidence.
\end{itemize}

\paragraph{(ii) Structured Retrieval-Augmented Generation.}
This family organizes memory into explicit graph structures to support
relational retrieval.
\begin{itemize}
\item \textbf{GraphRAG}~\cite{edge2024local} constructs an entity--relation
knowledge graph from the corpus and performs community-based, query-focused
summarization over it.
\item \textbf{HippoRAG}~\cite{gutierrez2024hipporag} is a neurobiologically
inspired method that builds a knowledge graph and uses a Personalized PageRank
scheme to emulate hippocampal indexing for multi-hop retrieval.
\item \textbf{LightRAG}~\cite{guo-etal-2025-lightrag} integrates graph
structures into indexing with a dual-level (low- and high-level) retrieval
process and incremental updates for efficiency.
\item \textbf{Youtu-GraphRAG}~\cite{dong2026youtu} employs vertically
unified agents over a graph for retrieval-augmented complex reasoning.
\end{itemize}
These methods reason over relational structure but rely on a topology that is
\emph{pre-built} and largely query-agnostic, which motivates our
query-conditioned latent graph.

\paragraph{(iii) Agentic Memory Management.}
This family maintains and updates an external memory bank with explicit
read/write operations.
\begin{itemize}
\item \textbf{MemoryBank}~\cite{zhong2024memorybank} augments the model with a
long-term memory store and an Ebbinghaus-inspired forgetting/updating mechanism.
\item \textbf{Mem0}~\cite{chhikara2025mem0} is a production-oriented memory layer
that extracts, consolidates, and retrieves salient facts for scalable long-term
memory.
\item \textbf{A-Mem}~\cite{xu2025mem} is an agentic memory that dynamically
organizes memories into interconnected notes with links and evolving structure.
\item \textbf{MemoryOS}~\cite{kang-etal-2025-memory} treats memory as an
operating-system-style hierarchy with scheduling across short-, mid-, and
long-term stores.
\item \textbf{LightMem}~\cite{fang2026lightmem} is a lightweight and efficient
memory-augmented generation framework that reduces the overhead of maintaining
an external memory.
\end{itemize}
These methods offer flexible read/write control but store memory as
\emph{textual} records, which are re-read as raw context and remain entangled
and noisy.

\paragraph{(iv) Reinforcement Learning for Memory.}
This family learns explicit memory-update policies via reinforcement learning.
\begin{itemize}
\item \textbf{Mem-$\alpha$}~\cite{wang2025mem} learns memory construction through
reinforcement learning, optimizing what to write into memory.
\item \textbf{MemAgent}~\cite{yu2026memagent} reshapes long-context processing
with a multi-conversation, RL-trained memory agent.
\item \textbf{Memory-R1}~\cite{yan2026memory} trains an agent to manage and
utilize memories (add / delete / update / retrieve) via reinforcement learning.
\item \textbf{MemCoE}~\cite{xu2026learning} is a cognition-inspired two-stage
optimization method that learns \emph{how} and \emph{what} to memorize for an
evolving memory; it is the strongest baseline in our main results.
\end{itemize}
These methods learn effective update policies but still operate over textual
memory and incur repeated prompting/updates, leading to higher inference cost.

% =====================================================================
\section{Algorithm}
\label{app:algorithm}
% =====================================================================

Algorithm~\ref{alg:lgm} gives a high-level, step-by-step description of
\texttt{LGM}. The procedure has three intuitive stages: (1) turn every past
interaction and the current query into a compact latent memory node;
(2) build a small graph that links the query to the memories most relevant to
it, and pass messages over this graph to obtain a single memory state; and
(3) feed that state to the language model as a few prefix tokens to generate the
answer. During training we additionally ask the same prefix to reconstruct the
supporting evidence, which prevents the memory state from collapsing into an
uninformative signal.

\begin{algorithm}[!t]
\caption{\texttt{LGM} (one forward pass)}
\label{alg:lgm}
\begin{algorithmic}[1]
\REQUIRE user history $\mathcal{H}=\{h_1,\dots,h_M\}$, current query $q$
\ENSURE personalized response $y$
\STATE \textbf{// Stage 1: encode into latent memory nodes}
\FOR{each item $x$ in $\mathcal{H}\cup\{q\}$}
    \STATE $z_x \leftarrow \textsc{Encode}(x)$
           \hfill\COMMENT{LLM hidden states $\rightarrow$ compact latent vector}
    \STATE $c_x \leftarrow \textsc{SAE}(z_x)$
           \hfill\COMMENT{disentangle $z_x$ into sparse concepts}
\ENDFOR
\STATE obtain query node $z_q$ and memory nodes $\{z_i\}_{i=1}^{M}$
\STATE
\STATE \textbf{// Stage 2: build a query-specific graph and propagate}
\FOR{each memory node $i$}
    \STATE $w_i \leftarrow \textsc{ConceptOverlap}(c_q, c_i)$
           \hfill\COMMENT{edge weight: relevance of memory $i$ to the query}
\ENDFOR
\STATE connect $q \leftrightarrow i$ for all memories with weights $\{w_i\}$
\FOR{$t = 1$ \TO $T$}
    \STATE every node updates itself by aggregating
           \STATE\hspace{1em} weighted messages from its neighbors
           \hfill\COMMENT{message passing}
\ENDFOR
\STATE $g \leftarrow \textsc{Readout}(\text{graph})$
       \hfill\COMMENT{one compact memory state for this query}
\STATE
\STATE \textbf{// Stage 3: generate the answer}
\STATE $P \leftarrow \textsc{Project}(g)$
       \hfill\COMMENT{turn $g$ into a few latent prefix tokens}
\STATE $y \leftarrow \textsc{LLM}([\,P;\,q\,])$
       \hfill\COMMENT{prepend prefix, then generate}
\RETURN $y$
\end{algorithmic}
\end{algorithm}

\paragraph{Training objective.}
The whole system is trained end-to-end by combining three terms: a
\emph{generation} loss that supervises the response $y^\star$, an
\emph{evidence-reconstruction} loss that forces the latent prefix to decode back
the supporting evidence $e^\star$ (so the memory state actually carries
information), and the SAE loss that keeps concepts sparse:
\begin{equation}
  \mathcal{L} = \mathcal{L}_{\text{gen}}
              + \mathcal{L}_{\text{recon}}
              + \mathcal{L}_{\text{sae}} .
\end{equation}

% =====================================================================
\section{Implementation and Hyperparameter Details}
\label{app:hyper}
% =====================================================================

Table~\ref{tab:hyperparams} lists the complete set of hyperparameters used to
train and evaluate \texttt{LGM}. All backbone / graph modules are optimized
jointly with AdamW~\cite{loshchilov2017decoupled} under bf16 mixed precision with
gradient checkpointing, and the random seed is fixed for reproducibility. By
default the backbone is fully fine-tuned; we additionally support a
parameter-efficient LoRA variant whose settings are listed at the bottom of the
table. Long histories are encoded in chunks (each segment truncated to $2048$
tokens) so that memory grows in the number of latent nodes rather than in a
monolithic context window.

\begin{table}[!t]
\centering
\footnotesize
\setlength{\tabcolsep}{4pt}
\renewcommand{\arraystretch}{0.98}
\begin{tabular}{ll}
\toprule
\textbf{Hyperparameter} & \textbf{Value} \\
\midrule
\multicolumn{2}{l}{\textit{Latent Encoding}} \\
Read-out layers $\mathcal{L}$        & $\{8, 16, 24, 32\}$ \\
Layer aggregation                    & mean \\
Sequence pooling                     & masked mean \\
Latent dimension $d$                 & $32$ \\
Chunk size (tokens)                  & $2048$ \\
\midrule
\multicolumn{2}{l}{\textit{Sparse Autoencoder (SAE)}} \\
Concept dimension                    & $256$ \\
Top-$k$ sparsity                     & $32$ \\
Reconstruction weight                & $0.1$ \\
$L_1$ sparsity weight                & $0.001$ \\
\midrule
\multicolumn{2}{l}{\textit{Relational Graph Network}} \\
Hidden size                          & $256$ \\
Message-passing steps $T$            & $4$ \\
Dropout                              & $0.1$ \\
\midrule
\multicolumn{2}{l}{\textit{Latent Prefix}} \\
\# prefix tokens $p$                 & $16$ \\
Norm rescaling                       & token-embedding norm \\
\midrule
\multicolumn{2}{l}{\textit{Optimization}} \\
Optimizer                            & AdamW \\
LR (graph modules)                   & $2\text{e-}4$ \\
LR (backbone)                        & $1\text{e-}5$ \\
Weight decay                         & $0.01$ \\
Gradient clipping                    & $1.0$ \\
Gradient accumulation                & $1$ \\
Epochs                               & $3$ \\
Precision                            & bf16 mixed \\
Gradient checkpointing               & enabled \\
Random seed                          & $23$ \\
\midrule
\multicolumn{2}{l}{\textit{Objective Weights}} \\
Generation loss                      & $0.05$ \\
Evidence reconstruction loss         & $0.35$ \\
Recon. target max tokens             & $96$ \\
\midrule
\multicolumn{2}{l}{\textit{Hardware \& Backbones}} \\
GPUs                                 & $8\times$ NVIDIA H800 \\
Max encode / LM tokens               & $2048$ / $2048$ \\
Backbones                            & \makecell[l]{Qwen2.5-7B-Instruct,\\ Gemma3-4B-it, Qwen3-4B-Instruct} \\
\midrule
\multicolumn{2}{l}{\textit{LoRA Variant (optional)}} \\
Rank $r$ / $\alpha$ / dropout        & $16$ / $32$ / $0.05$ \\
Target modules                       & \makecell[l]{q/k/v/o\_proj,\\ gate/up/down\_proj} \\
\bottomrule
\end{tabular}
\caption{Complete hyperparameter configuration of \texttt{LGM}.}
\label{tab:hyperparams}
\end{table}

\end{document}